\RequirePackage[T1]{fontenc}
\documentclass[conference,letterpaper]{IEEEtran}
\usepackage{amsmath,amssymb}
\usepackage{graphicx}
\usepackage{cite}
\usepackage[hidelinks]{hyperref}
\usepackage{url}
\usepackage{capt-of}
\usepackage{array}
\usepackage{booktabs}
\usepackage{multirow}
\usepackage[table]{xcolor}
\definecolor{codeorange}{HTML}{C06000}
\newcommand{\comparisonfont}{\rmfamily}
\title{AGQ: Action-Guided Ultra-Low-Bit Quantization for Vision-Language-Action Models}

\author{
\IEEEauthorblockN{\large
Xuhong Qiu\textsuperscript{1},
Zhuoyu Wu\textsuperscript{2,*},
Qiawei Zheng\textsuperscript{3},
Wenhui Ou\textsuperscript{3},
Wenqi Fang\textsuperscript{4},
Rapha\"{e}l C.-W. Phan\textsuperscript{2}
}

\IEEEauthorblockA{
\textsuperscript{1}
School of Computer Science and Engineering, Southeast University,
Nanjing, China
}

\IEEEauthorblockA{
\textsuperscript{2}
School of Information Technology,
Monash University,
Subang Jaya, Malaysia
}

\IEEEauthorblockA{
\textsuperscript{3}
Department of Electronic \& Computer Engineering,
Hong Kong University of Science \& Technology,
Hong Kong, China
}

\IEEEauthorblockA{
\textsuperscript{4}
Shenzhen Institutes of Advanced Technology, Chinese Academy of Sciences, Shenzhen, China}

\IEEEauthorblockA{
\textsuperscript{*}Corresponding author:
zhuoyu.wu@monash.edu
}
}

\begin{document}
\raggedbottom
\maketitle
\suppressfloats[t]
\begin{abstract}
Vision-language-action (VLA) models enable robots to perform tasks
specified through natural-language instructions, but their large
weight footprint challenges deployment on memory-constrained onboard
platforms. Ultra-low-bit quantization reduces this footprint, yet
preserving task performance and controlling weight recovery costs
remain challenging. We propose Action-Guided Quantization,
which combines action sensitivity with a measured proxy for weight
recovery cost to allocate matrix bitwidths under storage budgets.
Per-channel calibration compensates for quantization error, while
a uniform bitwidth within each matrix preserves regular layouts.
Bit-dependent grouped lossless coding reduces resident weight
storage, and block-wise recovery limits temporary buffers.
On LIBERO, AGQ achieves 92.7\% task success at W3,
exceeding AWQ and GPTQ by 1.2 and 3.7 percentage points,
respectively, and retains 70.5\% success at W2.5.
An asymmetric configuration, denoted W1.5$^{*}$, uses W4
vision weights and a parameter-weighted average of 1.5 bits
for language weights. With lossless coding, it achieves
6.10$\times$ complete-model compression relative to BF16
while retaining 25.5\% task success.
Code and experimental results are available at
\href{https://github.com/Qiuxu00/AGQ}{%
  \textcolor{codeorange}{\nolinkurl{https://github.com/Qiuxu00/AGQ}}}.
\end{abstract}

\begin{figure}[t]
\centering
\includegraphics[width=\columnwidth]{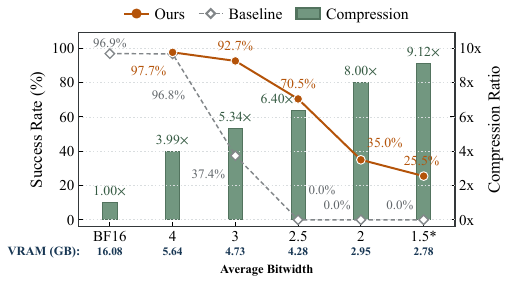}
\vspace{-8mm}
\caption{Low-bit quantization of VLA models.
Lines show task success rates (left axis) for Ours and Baseline (QVLA),
while bars show target-weight compression ratios relative to BF16
(right axis). QVLA reports 0\% at W2.5; lower-bit results are 
zero-valued extensions for visualization. 
W1.5*: W4 vision + W1.5 language.}
\label{fig:teaser}
\vspace{-3mm}
\end{figure}

\begin{figure*}[!t]
\centering
\includegraphics[trim=12bp 12bp 12bp 12bp,clip,width=1.0\textwidth]{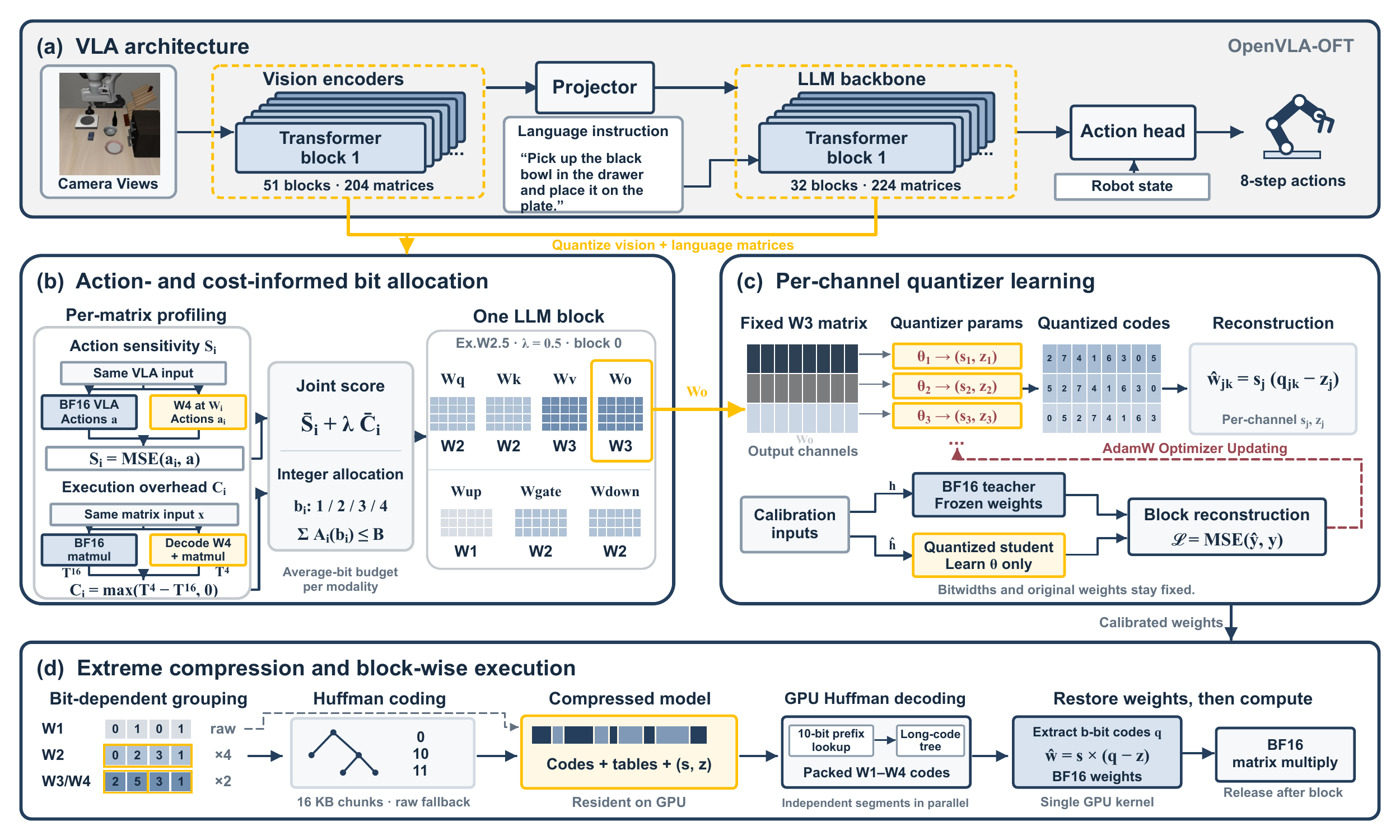}
\caption{Overview of AGQ:
(a) VLA architecture;
(b) action-guided bit allocation incorporating a measured
proxy for weight recovery cost;
(c) per-channel quantizer calibration; and
(d) grouped lossless coding and block-wise execution.}
\label{fig:framework}
\end{figure*}

\section{Introduction}

Vision-language-action (VLA) models integrate visual perception, language
understanding, and action prediction, enabling robots to perform tasks
specified through natural-language instructions~\cite{openvla, oft}.
Running these models locally is
attractive for robots that operate with limited connectivity or require
autonomous execution. However, their large weight footprint poses
a challenge for resource-constrained onboard computing platforms.
Ultra-low-bit weight quantization offers a way to reduce model
storage and make VLA inference feasible under tighter memory budgets.
These savings come at the risk of degraded action prediction:
in closed-loop control, quantization-induced action errors can
affect subsequent observations and task outcomes. Therefore, preserving task performance under aggressive compression remains a
central challenge.

Post-training quantization (PTQ) provides a practical approach to
compressing pretrained models using a small calibration
set~\cite{nagel}. AWQ uses activation statistics to protect salient
weights~\cite{awq}, while GPTQ uses second-order information to
compensate for quantization error~\cite{gptq}. For VLA models,
QVLA and ActQuant further incorporate action-related information
into quantization~\cite{qvla,actquant}. Beyond calibrating quantized
weights, mixed precision allows different layers or matrices to
use different bitwidths according to their sensitivity~\cite{mixedint,hawq}.
Hardware-aware approaches such as HAQ additionally incorporate
latency and energy feedback into precision selection~\cite{haq}.
Recent low-bit schemes also use structured mixed precision~\cite{huang2024slim},
lattice codebooks~\cite{tseng2024quip}, and dense-and-sparse
quantization~\cite{kim2023squeezellm}.

Ultra-low-bit VLA deployment requires allocating limited precision
across matrices with different effects on action prediction.
Our analysis reveals sensitivity differences even within the same
layer, motivating matrix-wise allocation. The deployment cost also
extends beyond average bitwidth: metadata and padding increase
storage, while weight recovery requires computation and temporary
buffers. Entropy-based weight compression supports memory-efficient
deployment~\cite{hao2024neuzip,zhang202670}, while recovered BF16 weights
still require temporary buffers. 

To address these issues, we propose \textbf{Action-Guided
Quantization} (AGQ). AGQ selects a bitwidth for each weight matrix
based on action sensitivity and a measured estimate of weight
recovery cost under storage constraints. Per-channel calibration
reduces quantization error while keeping a uniform bitwidth within
each matrix. Grouped lossless coding reduces weight storage, and
block-wise recovery reduces temporary buffer requirements.
As shown in Fig.~\ref{fig:teaser}, AGQ achieves a better tradeoff
between task success and target-weight compression than the
baseline quantization method.

In summary, we make the following contributions:
\begin{itemize}
    \item We design Action-Guided Quantization (AGQ), a matrix-wise
    allocation method that combines an action-sensitivity prior
    from a common W4 probe with measured weight recovery cost
    under explicit storage budgets.

    \item We propose a calibration and execution pipeline that
    combines per-channel quantizer learning, grouped lossless
    coding, and block-wise recovery to reduce quantization error
    and resident and temporary weight storage.

    \item We evaluate AGQ in closed-loop tasks on LIBERO and measure
    its deployment performance on an RTX 3090. AGQ achieves
    92.7\% and 70.5\% overall success at W3 and W2.5, respectively.
    An asymmetric W4-vision/W1.5-language configuration achieves
    25.5\% success with $6.10\times$ complete-model compression
    relative to BF16.
\end{itemize}
\section{Proposed Method}
\label{sec:method}
\subsection{Overview}
Fig.~\ref{fig:framework} separates preparation from compressed inference.
Starting from a BF16 policy, calibration observations, and backbone storage
budgets, stage (b) assigns matrix bitwidths using action sensitivity and
execution overhead. Stage (c) fixes this allocation and learns channel
quantizers through block reconstruction. Stage (d) encodes the resulting
integer weights for deployment. During inference, encoded weights remain
GPU-resident; decoding and fused unpacking/dequantization recover the active
block's BF16 weights, which are released on exit. Allocation and parameter learning remain offline.

\subsection{Action-Guided Bit Allocation}
We assign one bitwidth per matrix to preserve regular decoding. A common W4
perturbation provides a lower-bit allocation prior, assuming some consistency
in relative sensitivity across precisions. One probe per matrix replaces
separate sensitivity scans at every candidate bitwidth, reducing preparation
work without assuming identical rankings.

For matrix $i$, we quantize only that matrix to W4 and retain all other
parameters in BF16. Its sensitivity is the same-input action deviation:
\begin{equation}
S_i=\mathbb E_{\mathrm{bal}}\!\left[
\frac{1}{6H}\sum_{h=0}^{H-1}
\left\|\mathbf a_h^{(i)}-\mathbf a_h^{\mathrm{BF16}}\right\|_2^2
\right].
\label{eq:sensitivity}
\end{equation}
Each vector contains six normalized components: three for position and three
for rotation, excluding the gripper. With $H=8$ predicted actions, $6H$ averages
over all components and steps. Tasks and their trajectories receive equal
weight, preventing longer trajectories from dominating.

Same-input W4 timing provides a measured proxy for weight-recovery overhead.
We normalize sensitivity and cost separately within each backbone:
\begin{align}
C_i&=\max\!\left(T_i^{\mathrm{W4}}-T_i^{\mathrm{BF16}},0\right),
\label{eq:cost}\\
\rho_i&=\frac{S_i}{\sum_j S_j}+\lambda\frac{C_i}{\sum_j C_j}.
\label{eq:score}
\end{align}
Here $\lambda$ controls the overhead contribution. Under separate vision and
language budgets, we solve
\begin{equation}
\begin{aligned}
\min_{\{b_i\}}\quad&\sum_i\rho_i(4-b_i)^2,\\
\mathrm{s.t.}\quad&\sum_i A_i(b_i)\leq B,\quad b_i\in\{1,2,3,4\},
\end{aligned}
\label{eq:allocation}
\end{equation}
where $A_i(b_i)$ accounts for candidate-format storage, including metadata
and padding, and $B$ is the backbone budget. The quadratic penalty implements
a conservative allocation rule: matrices with greater action sensitivity or
measured recovery overhead receive stronger protection from aggressive
bitwidth reduction. W4 timing sets this priority; it does not predict latency
at each candidate precision. Reported compression ratios use actual exported bytes.

\subsection{Per-Channel Calibration for Regular Execution}
Per-channel scale $s_j$ sets the quantization step, and integer zero point
$z_j$ locates real zero. Sharing them limits metadata and preserves regular
access during dequantization.

Following OmniQuant's clipping and block reconstruction~\cite{omniquant}, we
learn clipping coefficients for channel bounds $[\ell_j,u_j]$. For W2--W4,
let $Q_i=2^{b_i}-1$:
\begin{gather}
s_j=\frac{u_j-\ell_j}{Q_i},\quad
z_j=\operatorname{clip}\!\left(\operatorname{round}(-\ell_j/s_j),0,Q_i\right).
\label{eq:channel_parameters}\\
q_{jk}=\operatorname{clip}\!\left(\operatorname{round}(W_{jk}/s_j)+z_j,0,Q_i\right).
\label{eq:weight_quantization}
\end{gather}
The codes reconstruct weights as
\mbox{$\widehat W_{jk}=s_j(q_{jk}-z_j)$}.
We floor $s_j$ at a positive constant and round it to BF16 before computing
$z_j$. Weights and bitwidths remain fixed; W1 learns channel amplitudes with
signs fixed.

Frozen BF16 block outputs provide reconstruction targets~\cite{li2021brecq}:
\begin{equation}
\mathcal L_\ell=\mathbb E_{u\in\mathcal D}
\operatorname{MSE}\!\left(
f_\ell(\widehat h_u;\widehat W_\ell),f_\ell(h_u;W_\ell)
\right).
\label{eq:reconstruction}
\end{equation}
Here $h_u$ is the teacher input and $\widehat h_u$ comes from preceding quantized
blocks. AdamW updates the clipping coefficients by backpropagation; the resulting channel
bounds determine $s_j$ and $z_j$, which regenerate quantized weights. Each
configuration starts independently from BF16, with a larger calibration share
for long-horizon tasks.

Deployment uses these parameters in a GPU kernel that fuses unpacking and
per-channel dequantization, followed by BF16 matrix multiplication.

\subsection{Bit-Dependent Grouped Coding and Execution}
Entropy coding further compresses quantized weights~\cite{han2015deep}.
Our bit-dependent grouping balances codebook size and values recovered per symbol.

Four adjacent W2 values or two W3/W4 values form a joint Huffman symbol~\cite{huffman}, while W1 retains raw packing. These six- or eight-bit symbols capture joint distributions with at most 256 symbols, recovering multiple values per decode operation.

To reduce decoding dependencies~\cite{sitaridi2016massively}, we partition
fixed-size coding blocks into independently decodable segments and use
prefix lookup to reduce bit-by-bit traversal. Per-matrix encoding selects the smaller representation, including codebooks and block indexes, with raw packing as fallback. Scales and zero points remain unchanged, ensuring no additional quantization error from lossless decoding.

Encoded weights and metadata remain GPU-resident. On block entry, decoding restores packed codes for fused unpacking/dequantization. Temporary BF16 weights for seven language or four vision matrices are released on
block exit, reducing resident storage and bounding temporary dense weights by the largest active block.

\begingroup\brokenpenalty=10000
\par\noindent\begin{minipage}{\columnwidth}
\vspace*{5pt}
\section{Evaluation}
\label{sec:evaluation}
\subsection{Experimental Setup}
Following ActQuant~\cite{actquant}, we evaluate OpenVLA-OFT on four LIBERO
suites~\cite{libero}. Some weights are calibrated on A100; all evaluations
use RTX3090. Ten configurations each run 600 closed-loop episodes
(150 per suite), totaling 6,000. Metrics include overall and per-suite
success, complete-model storage, peak GPU memory, and mean latency over
ten fixed inputs. Each call predicts eight actions; timing excludes
observation preprocessing and action postprocessing.
\end{minipage}\par

\subsection{Effectiveness of Joint Bit Allocation}
\label{sec:allocation_effectiveness}
\begin{table}[!tb]
\caption{Matrix-wise bit allocation with supporting probe analysis}
\label{tab:allocation_analysis}
\centering
\begingroup
\fontsize{7.7}{9.4}\selectfont
\setlength{\tabcolsep}{1.3pt}
\renewcommand{\arraystretch}{1.0}
\setlength{\aboverulesep}{1.4pt}
\setlength{\belowrulesep}{1.4pt}
\arrayrulecolor{black!65}
\setlength{\arrayrulewidth}{0.25pt}
\begin{tabular*}{\columnwidth}{@{\extracolsep{\fill}}c|c|cccc}
\toprule[0.8pt]
\multirow{2}{*}{\textbf{Budget}} & \multirow{2}{*}{$\boldsymbol{\lambda}$} & \multicolumn{4}{c}{\textbf{Total (428)}} \\
\cmidrule(lr){3-6}
 & & \textbf{W1} & \textbf{W2} & \textbf{W3} & \textbf{W4} \\
\midrule[0.5pt]
\multirow{5}{*}{\textbf{W2.5}} & 0 & 54 (23+31) & 107 (67+40) & 189 (63+126) & 78 (51+27) \\
 & 0.25 & 28 (8+20) & 142 (91+51) & 192 (55+137) & 66 (50+16) \\
 & 0.5 & 19 (6+13) & 150 (94+56) & 199 (51+148) & 60 (53+7) \\
 & 0.75 & 14 (5+9) & 155 (96+59) & 199 (45+154) & 60 (58+2) \\
 & 1 & 7 (5+2) & 167 (95+72) & 196 (47+149) & 58 (57+1) \\
\midrule[0.5pt]
\multirow{5}{*}{\textbf{W2}} & 0 & 133 (69+64) & 111 (42+69) & 138 (57+81) & 46 (36+10) \\
 & 0.25 & 107 (60+47) & 139 (54+85) & 149 (59+90) & 33 (31+2) \\
 & 0.5 & 95 (55+40) & 159 (61+98) & 146 (61+85) & 28 (27+1) \\
 & 0.75 & 88 (50+38) & 164 (69+95) & 150 (60+90) & 26 (25+1) \\
 & 1 & 86 (49+37) & 169 (72+97) & 142 (53+89) & 31 (30+1) \\
\bottomrule[0.8pt]
\end{tabular*}
\endgroup
\par\smallskip
\includegraphics[width=\columnwidth]{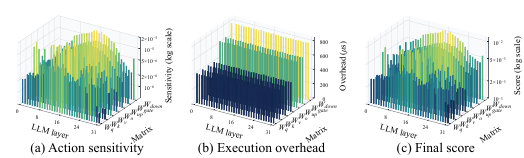}
\par\smallskip
\footnotesize\raggedright
Counts: total (vision + language). Panels: language W4 probes at W2.5,
$\lambda=0.5$; (a,c) logarithmic axes, (b) microseconds.
\end{table}

Table~\ref{tab:allocation_analysis} reports allocations and W4 probes. Action sensitivity spans $20.2\times$
and recovery overhead $2.0\times$, with differences within layers supporting
matrix-wise allocation. At W2.5, increasing $\lambda$ from 0 to 0.5 reduces
W1 matrices from 54 to 19, yielding W1--W4 counts of 19, 150, 199, and 60
under the storage budget.

\par\addvspace{10pt}\noindent
\begin{minipage}{\columnwidth}
\centering
\includegraphics[width=\columnwidth]{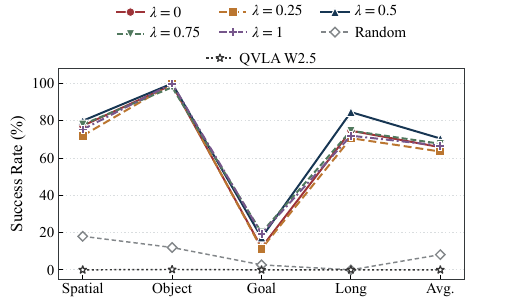}
\par\nobreak
\makeatletter\def\@captype{figure}\makeatother
\setlength{\abovecaptionskip}{4pt}
\caption{W2.5 success rates across four suites and overall.
Random allocation also undergoes calibration. QVLA values are from
ActQuant~\cite{actquant}.}
\label{fig:lambda_success}
\end{minipage}
\par\addvspace{10pt}

Figure~\ref{fig:lambda_success} compares guided allocation with calibrated
random allocation and QVLA at W2.5. Our five $\lambda$ settings achieve
63.5--70.5\% overall success, versus 8.2\% for random allocation and 0.0\%
for QVLA. The best overall setting, $\lambda=0.5$, reaches 70.5\%, improving
on sensitivity-only allocation ($\lambda=0$) by 4.7 percentage points.
This advantage holds for the overall average, rather than every suite;
subsequent results use $\lambda=0.5$ unless stated otherwise.

\subsection{Task Success at Ultra-Low Bitwidths}
\label{sec:quality_comparison}
\begin{table}[t]
\caption{Closed-loop task success rates (\%) on LIBERO}
\label{tab:quality_comparison}
\centering
\definecolor{rowunderline}{HTML}{8B2437}
\newcommand{\redunder}[1]{\begingroup\setbox0=\hbox{#1}\leavevmode\rlap{\raisebox{-2.0pt}[0pt][0pt]{\color{rowunderline}\rule{\wd0}{0.65pt}}}\box0\endgroup}
\definecolor{bestpink}{HTML}{F7E1E6}
\begingroup\comparisonfont\fontsize{7.3}{8.9}\selectfont
\setlength{\tabcolsep}{1.3pt}
\renewcommand{\arraystretch}{1.0}
\setlength{\aboverulesep}{1.4pt}
\setlength{\belowrulesep}{1.4pt}
\arrayrulecolor{black!65}
\setlength{\arrayrulewidth}{0.25pt}

\begin{tabular*}{\linewidth}{@{\extracolsep{\fill}}c|c|c|ccccc|c}
\toprule[0.8pt]
\multirow{2}{*}{\textbf{Model}} & \multirow{2}{*}{\textbf{Method}} & \multirow{2}{*}{\shortstack{\textbf{Vision+}\\\textbf{LLM BPW}}} & \multicolumn{5}{c|}{\textbf{Success Rate (\%)}} & \\
\cmidrule(lr){4-8}
 & & & \textbf{Spatial} & \textbf{Object} & \textbf{Goal} & \textbf{Long} & \textbf{Avg.} & \textbf{$\Delta$} \\
\midrule[0.5pt]
 & Baseline & 16.0 & 97.6 & 98.4 & 96.8 & 95.1 & 96.9 & 0.0 \\
\cmidrule(l){2-9}
 &  & 4.0 & 94.6 & 98.6 & 96.8 & 94.4 & 96.1 & $-0.8$ \\
 &  & 3.0 & 86.4 & 98.0 & \cellcolor{bestpink}\textbf{90.2} & \cellcolor{bestpink}\textbf{91.4} & 91.5 & $-5.4$ \\
 & \multirow{-3}{*}{AWQ} & 2.0 & 0.0 & 0.0 & 0.0 & 0.0 & 0.0 & $-96.9$ \\
\cmidrule(l){2-9}
 &  & 4.0 & 90.8 & 98.8 & 96.4 & 92.6 & 94.6 & $-2.3$ \\
 &  & 3.0 & 82.4 & 98.0 & 88.8 & 87.2 & 89.0 & $-7.9$ \\
 & \multirow{-3}{*}{GPTQ} & 2.0 & 0.0 & 0.0 & 0.0 & 0.0 & 0.0 & $-96.9$ \\
\cmidrule(l){2-9}
 &  & 4.0 & 96.0 & 98.7 & 97.4 & \cellcolor{bestpink}\textbf{95.2} & 96.8 & $-0.1$ \\
 &  & 3.5 & 82.9 & 97.2 & 91.4 & 93.6 & 91.3 & $-5.6$ \\
 &  & 3.0 & 40.8 & 56.8 & 20.2 & 31.8 & 37.4 & $-59.5$ \\
 & \multirow{-4}{*}{QVLA} & 2.5 & 0.0 & 0.2 & 0.0 & 0.0 & 0.0 & $-96.9$ \\
\cmidrule(l){2-9}
\rowcolor{black!7}
\cellcolor{white} &  & 4.0 & \cellcolor{bestpink}\textbf{97.3} & \cellcolor{bestpink}\textbf{100.0} & \cellcolor{bestpink}\textbf{98.7} & 94.7 & \cellcolor{bestpink}\textbf{97.7} & \cellcolor{bestpink}\textbf{+0.8} \\
\rowcolor{black!7}
\cellcolor{white} &  & 3.0 & \cellcolor{bestpink}\textbf{96.7} & \cellcolor{bestpink}\textbf{100.0} & 83.3 & 90.7 & \cellcolor{bestpink}\textbf{92.7} & \cellcolor{bestpink}\textbf{$-4.2$} \\
\rowcolor{black!7}
\cellcolor{white} &  & 2.5 & \cellcolor{bestpink}\textbf{80.0} & \cellcolor{bestpink}\textbf{100.0} & \cellcolor{bestpink}\textbf{17.3} & \cellcolor{bestpink}\textbf{84.7} & \cellcolor{bestpink}\textbf{70.5} & \cellcolor{bestpink}\textbf{$-26.4$} \\
\rowcolor{black!7}
\cellcolor{white} &  & 2.0 & \cellcolor{bestpink}\textbf{29.3} & \cellcolor{bestpink}\textbf{92.0} & \cellcolor{bestpink}\textbf{1.3} & \cellcolor{bestpink}\textbf{17.3} & \cellcolor{bestpink}\textbf{35.0} & \cellcolor{bestpink}\textbf{$-61.9$} \\
\rowcolor{black!7}
\cellcolor{white}\multirow{-16}{*}[5pt]{\shortstack[c]{\textbf{OpenVLA-}\\\textbf{OFT}}} & \multirow{-5}{*}{\textbf{Ours}} & \redunder{1.5\textsuperscript{*}} & \redunder{38.7} & \redunder{60.7} & \redunder{0.0} & \redunder{2.7} & \redunder{25.5} & \redunder{$-71.4$} \\
\bottomrule[0.8pt]
\end{tabular*}
\endgroup
\par\smallskip
\footnotesize\raggedright
AWQ and GPTQ use uniform precision; QVLA and Ours use mixed
precision (average BPW). Bold entries with pale pink shading indicate the
best results at a shared bitwidth. Dark-red underlines emphasize
W1.5$^{*}$, for which no same-bitwidth baseline is available.
$\Delta$ denotes the average success-rate difference from Baseline
in percentage points. $^{*}$W4 vision + W1.5 language.
\end{table}

Table~\ref{tab:quality_comparison} compares task success across precisions.
At W3, AGQ reaches 92.7\%, exceeding AWQ, GPTQ, and QVLA by 1.2, 3.7,
and 55.3 percentage points. It retains 70.5\% at W2.5 versus QVLA's 0.0\%,
and 35.0\% at W2 versus 0.0\% for AWQ and GPTQ. W1.5$^{*}$ combines W4
vision with W1.5 language weights (approximately 1.75 average bits across
targeted weights) and retains 25.5\% success. At W4, AGQ achieves 97.7\%
versus 96.9\% for BF16.

\subsection{Storage and Inference Efficiency}
\label{sec:performance}
\begin{table}[t]
\caption{Storage and inference efficiency of OpenVLA-OFT}
\label{tab:performance_comparison}
\centering
\begingroup\fontsize{7.7}{9.4}\selectfont
\setlength{\tabcolsep}{1.1pt}
\renewcommand{\arraystretch}{1.05}
\setlength{\aboverulesep}{1.8pt}
\setlength{\belowrulesep}{1.8pt}
\arrayrulecolor{black!65}
\setlength{\arrayrulewidth}{0.25pt}

\begin{tabular*}{\linewidth}{@{\extracolsep{\fill}}c|c|c|ccc}
\toprule[0.8pt]
\textbf{Model} & \textbf{Method} & \shortstack{\textbf{Vision+}\\\textbf{LLM BPW}} & \shortstack{\textbf{Model Size}\\\textbf{(GB)}} & \shortstack{\textbf{Peak GPU}\\\textbf{Mem. (GB)}} & \shortstack{\textbf{Latency}\\\textbf{(ms)}} \\
\midrule[0.5pt]
 & BF16 & 16.0 & 15.457 (1.00$\times$) & 16.082 & 187.3 \\
\cmidrule(l){2-6}
\rowcolor{black!7}
\cellcolor{white} &  & 4.0 & 4.680 (3.30$\times$) & 5.643 & 249.3 \\
\rowcolor{black!7}
\cellcolor{white} &  & 3.0 & 3.774 (4.10$\times$) & 4.727 & 254.4 \\
\rowcolor{black!7}
\cellcolor{white} &  & 2.5 & 3.327 (4.65$\times$) & 4.280 & 245.2 \\
\rowcolor{black!7}
\cellcolor{white} &  & 2.0 & 2.878 (5.37$\times$) & 3.828 & 243.6 \\
\rowcolor{black!7}
\cellcolor{white} & \multirow{-5}{*}{\textbf{Ours}} & 1.5$^{*}$ & 2.656 (5.82$\times$) & 3.603 & 250.5 \\
\cmidrule(l){2-6}
\rowcolor{black!7}
\cellcolor{white} &  & 2.5 & 3.123 (4.95$\times$) & 4.136 & 473.9 \\
\rowcolor{black!7}
\cellcolor{white}\multirow{-8}{*}[3pt]{\shortstack[c]{\textbf{OpenVLA-}\\\textbf{OFT}}} & \multirow{-2}{*}{\shortstack{\textbf{Ours} +\\Lossless Coding}} & 1.5$^{*}$ & 2.534 (6.10$\times$) & 3.537 & 310.4 \\
\bottomrule[0.8pt]
\end{tabular*}
\endgroup
\par\smallskip
\footnotesize\raggedright
Parentheses show complete-model compression relative to BF16.
BPW denotes vision + language weight precision (average for Ours).
$^{*}$W4 vision + W1.5 language.
\end{table}

Table~\ref{tab:performance_comparison} reports deployment efficiency.
\textbf{Memory.} At W1.5$^{*}$, quantization reduces complete-model storage
from 15.457 to 2.656\,GB; lossless coding further reduces it to 2.534\,GB
($6.10\times$ compression). Peak GPU memory falls from 16.082 to 3.537\,GB
(78.0\%), as block-wise recovery bounds temporary BF16 weights.
Ratios include all model parameters and quantization metadata.

\textbf{Execution.} Uncoded W4--W1.5$^{*}$ configurations take
243.6--254.4\,ms versus 187.3\,ms for BF16. This narrow latency range is
consistent with execution dominated by recovery and BF16 matrix
multiplication rather than packed weight size.

\textbf{Storage--latency trade-off.} Coding raises W1.5$^{*}$ latency
from 250.5 to 310.4\,ms and W2.5 latency from 245.2 to 473.9\,ms,
showing configuration-dependent overhead. W1.5$^{*}$ offers the smallest
footprint at 310.4\,ms and 25.5\% success; greater storage compression
therefore does not necessarily accelerate inference. Coding saves
0.122\,GB at W1.5$^{*}$ for 59.9\,ms additional latency, compared with
0.204\,GB and 228.7\,ms at W2.5. Storage gains must therefore be assessed
jointly with decoding cost.

Published OpenVLA-OFT times on RTX4090 range from 152.6\,ms (NF4) to
379.5\,ms (LLM.int8())~\cite{vlaquantbench}. These LLM-only rollout
measurements provide context rather than a matched speedup comparison.
\par\endgroup

\section{Conclusion}
We presented Action-Guided Quantization (AGQ) for ultra-low-bit
VLA deployment. AGQ uses action sensitivity and a measured proxy
for weight recovery cost to guide matrix-wise bit allocation,
complemented by per-channel calibration, grouped lossless coding,
and block-wise recovery. On LIBERO, AGQ outperforms the compared
baselines at shared precisions below four bits and achieves
70.5\% task success at W2.5. The W1.5$^{*}$ configuration achieves
$6.10\times$ complete-model compression while retaining 25.5\%
success.

\makeatletter
\setlength{\@fptop}{0pt}
\setlength{\@fpsep}{9pt}
\setlength{\@fpbot}{0pt plus 1fil}
\makeatother
\clearpage
\bibliographystyle{IEEEtran}
\bibliography{references}
\end{document}